\documentclass[sigconf]{acmart}

\usepackage{verbatim}

\usepackage{float}
\usepackage[utf8]{inputenc} 
\usepackage[T1]{fontenc}    
\usepackage{hyperref}       
\usepackage{url}            
\usepackage{booktabs}       
\usepackage{amsfonts}       
\usepackage{nicefrac}       
\usepackage{microtype}      
\usepackage{lipsum}		
\usepackage{graphicx}
\usepackage{natbib}
\usepackage{doi}
\usepackage{tikz}
\usepackage{pgfplots}
\usepackage{caption}
\usepackage{subcaption}
\usepackage{amsmath} 
\usepackage{multirow}
\usepackage{enumitem}
\usepackage{svg}
\usepackage{soul}
\usepackage{bbm}
\usepackage{hyperref}
\usepackage[mathbold]{mathfixs}
\hypersetup{
    colorlinks=true,
    linkcolor=blue,
    filecolor=magenta,      
    urlcolor=cyan,
    pdftitle={Overleaf Example},
    pdfpagemode=FullScreen,
    }

\AtBeginDocument{%
  }
\pgfplotsset{compat=1.18}
\setcopyright{acmlicensed}
\copyrightyear{2018}
\acmYear{2018}
\acmDOI{XXXXXXX.XXXXXXX}

\acmConference[CIKM 2025]{The Conference on Information and Knowledge Management}{Nov. 10--Nov. 14, 2025}{Seoul, Korea}
\acmISBN{978-1-4503-XXXX-X/18/06}

\begin{document}


\title{CREMA: A Contrastive Regularized Masked Autoencoder for Robust ECG Diagnostics across Clinical Domains}

\author{Junho Song}
\email{jhsong@medicalai.com}
\affiliation{
  \institution{AI Group, Medical AI Co., Ltd.}
  \city{Seoul}
  \country{Republic of Korea}
}
\author{Jong-Hwan Jang}
\email{jangood1122@medicalai.com}
\affiliation{
  \institution{AI Group, Medical AI Co., Ltd.}
  \city{Seoul}
  \country{Republic of Korea}
}
\author{DongGyun Hong}
\email{dghong@medicalai.com}
\affiliation{
  \institution{AI Group, Medical AI Co., Ltd.}
  \city{Seoul}
  \country{Republic of Korea}
}
\author{Joon-myoung Kwon}
\email{cto@medicalai.com}
\affiliation{
  \institution{AI Group, Medical AI Co., Ltd.}
  \city{Seoul}
  \country{Republic of Korea}
}
\author{Yong-Yeon Jo}
\email{yy.jo@medicalai.com}
\affiliation{
  \institution{AI Group, Medical AI Co., Ltd.}
  \city{Seoul}
  \country{Republic of Korea}
}

\renewcommand{\shortauthors}{Junho Song et al}

\begin{abstract}
Electrocardiogram (ECG) diagnosis remains challenging due to limited labeled data and the need to capture subtle yet clinically meaningful variations in rhythm and morphology. We present CREMA (Contrastive Regularized Masked Autoencoder), a foundation model for 12-lead ECGs designed to learn generalizable representations through self-supervised pretraining. CREMA combines generative learning and contrastive regularization via a Contrastive Regularized MAE loss, and employs a Signal Transformer (SiT) architecture to capture both local waveform details and global temporal dependencies. We evaluate CREMA on benchmark datasets and real-world clinical environments, including deployment scenarios with significant distribution shifts. CREMA outperforms supervised baselines and existing self-supervised models in both linear probing and fine-tuning evaluations. Notably, it maintains superior performance across diverse clinical domains, such as emergency care, highlighting its robustness under real-world conditions. These results demonstrate that CREMA serves as a scalable and reliable foundation model for ECG diagnostics, supporting downstream applications across heterogeneous and high-risk clinical settings.
\end{abstract}

\ccsdesc[500]{Applied computing~Health care information systems}

\keywords{Foundation model, Self-supervised learning, Generative learning, Contrastive learning, Electrocardiogram, Biosignal process}


\maketitle

\section{Introduction}

Electrocardiograms (ECGs) are time-series recordings of the heart’s electrical activity, capturing information on rhythm, strength, timing, and beat regularity~\cite{hinton2018incidence}. These signals are essential for detecting cardiac conditions such as myocardial infarction (MI), arrhythmias, and other structural or functional abnormalities. As such, accurate modeling of ECGs plays a critical role in enabling timely and precise diagnosis.

Supervised learning has shown strong performance in ECG tasks such as classification, prediction, and denoising. However, these models rely heavily on large, labeled datasets, which are often scarce due to privacy concerns and the low prevalence of many cardiac disorders~\cite{hinton2018incidence}. This scarcity frequently results in class imbalance and hinders generalizable model training.

To mitigate these issues, recent work has turned to self-supervised learning (SSL) applied to large-scale unlabeled ECG datasets~\cite{devlin2018bert, achiam2023gpt, chen2020simple, he2022masked, baevski2020wav2vec, kiyasseh2020clocs, wang2023contrast, na2024guiding}. SSL-based pretraining enables models to learn general representations that transfer well to downstream tasks, offering three major advantages: (1) \textit{robust feature extraction across domains}, (2) \textit{improved fine-tuning performance compared to training from scratch}, and (3) \textit{faster convergence with limited labeled data}.

Despite this promise, applying SSL to ECGs remains challenging. Unlike generic time series, ECGs contain subtle but clinically meaningful variations in waveform shape, intervals, and rhythm that require fine-grained modeling~\cite{zheng202012, shekatkar2017detecting}. To capture these patterns, both contrastive and generative learning approaches have been explored~\cite{lai2023practical, wang2024contrast, chien2022maeeg, na2024guiding}. However, contrastive learning methods often use ECG-specific augmentations, such as cutout and dropout, that risk \textit{distorting ECG diagnostic information}~\cite{na2024guiding, liu2024zero}. In contrast, generative learning methods that rely on reconstruction objectives tend to produce \textit{overly dense embeddings, limiting discriminability}~\cite{radhakrishnan2023cross}.

To address these limitations, we introduce \textbf{CREMA} (Contrastive Regularized Masked Autoencoder), a foundation model for 12-lead ECG diagnostics. CREMA builds on the \textbf{Signal Transformer (SiT)} architecture, which combines a 1D convolution block with a Vision Transformer (ViT)~\cite{dosovitskiy2020image}. While ViT effectively models long-range temporal dependencies, it lacks explicit mechanisms for capturing localized waveform features, such as P waves, QRS complexes, and T waves, essential for ECG interpretation. The added convolution block addresses this limitation by extracting local morphology before passing patch embeddings to the transformer. This design enables CREMA to represent both fine-grained and global patterns in ECGs more effectively.CREMA is trained using both generative learning (GL) and contrastive learning (CL), unified through a novel \textbf{Contrastive Regularized MAE Loss} that encourages both reconstruction fidelity and representation separability. This design allows CREMA to extract both local morphological features and global rhythm patterns, resulting in robust and generalizable ECG representations.

In extensive experiments, CREMA outperforms other SSL-based models in both linear probing and fine-tuning scenarios. The contrastive regularization further accelerates convergence and improves discriminability by mitigating the embedding density commonly seen in pure generative models. When deployed in our real-world ECG analysis service (AiTiA-Series, \url{https://aitia-demo.medicalai.com}), diagnostic models fine-tuned on CREMA surpass legacy supervised baselines. Moreover, under distribution shift settings, training on one institution and testing on another, CREMA consistently achieves superior performance, demonstrating robustness across diverse clinical domains. 

\vspace{0.2cm}
\noindent Our contributions are summarized as follows:

\begin{itemize}[leftmargin=*]

    \item We present a Signal Transformer (SiT) that combines convolution and transformer layers for effective ECG representation learning.
    \item We introduce a contrastive-regularized MAE loss that balances generative and contrastive objectives.
    \item We present CREMA as a foundation model and validate its effectiveness through benchmark and clinical evaluations.
    \item We demonstrate CREMA’s superiority in real-world deployment settings, where it is actively used in live diagnostic services.
    \item We analyze the impact of contrastive regularization on robust ECG representation across domain shifts.
\end{itemize}

This study establishes the pivotal role of foundation models, exemplified by CREMA, in advancing the state of ECG diagnostics. By effectively addressing the complexities inherent to ECG data and leveraging a hybrid learning approach, CREMA serves as a benchmark for precision and scalability in clinical applications. 

\section{Related Work}
\subsection{SSL for ECG Representation Learning}

Recent advances in ECG analysis have highlighted the efficacy of SSL for foundation model development. SSL approaches are particularly promising in low-label settings, enabling the extraction of robust and transferable ECG representations. The primary SSL paradigms include contrastive learning (CL), generative learning (GL), and hybrid learning (HL) strategies~\cite{yue2022ts2vec, jiang2022transferability, franceschi2019unsupervised, nonnenmacher2022utilizing, sarkar2020self, Sarkar2020Self-Supervised, rebjock2021online, sun2021adjusting, yeche2021neighborhood, yang2022unsupervised, cheng2023timemae}.

\textbf{Contrastive Learning (CL)}:  
CL enhances representation discriminability by aligning augmented views of the same sample while separating views from different samples. Models such as SimCLR~\cite{chen2020simple}, CLoCS~\cite{kiyasseh2020clocs}, and COMET~\cite{wang2023contrast} have demonstrated promising results in ECG representation learning. However, standard augmentation methods used in CL, including cutout and dropout~\cite{na2024guiding}, may distort semantic integrity in ECGs~\cite{lan2023towards}.

\textbf{Generative Learning (GL)}:  
GL focuses on reconstructing masked portions of input data to learn fine-grained waveform structure. Approaches like MAE~\cite{he2022masked} and ST-MEM~\cite{na2024guiding} have shown promise in modeling ECG morphology. However, GL tends to produce dense embeddings due to its reconstruction-centric objective, which can hinder discriminability~\cite{radhakrishnan2023cross}.

\textbf{Hybrid Learning}:  
Recent studies have explored combining contrastive and generative learning to capture both discriminative and reconstructive features~\cite{huang2023contrastive}. For example, contrastive objectives are applied to MAE-encoded representations to improve separability while preserving signal fidelity. However, most hybrid models still rely on manually tuned loss weights and are evaluated on limited or homogeneous datasets~\cite{huang2023contrastive, yang2022unsupervised}, limiting their generalizability to diverse or real-world applications.

\subsection{Clinical Deployment Considerations}

As self-supervised ECG models advance toward clinical deployment, ensuring robustness to data heterogeneity, scalability, and integration efficiency becomes increasingly important. While many models achieve strong performance on curated benchmarks~\cite{na2024guiding, lee2022efficient}, few have been systematically evaluated under real-world distribution shifts, such as variations across institutions, devices, or patient populations. In addition, architectural efficiency and latency constraints essential for deployment in real-time or point-of-care settings are often overlooked.

These limitations highlight the need for ECG foundation models that not only generalize across diverse clinical environments but also maintain a balance between semantic fidelity and discriminative utility~\cite{radhakrishnan2023cross, mckeen2024ecg}. To address these gaps, we propose a contrastive-regularized generative pretraining strategy designed to enhance robustness under clinically realistic conditions.

\begin{figure*}[!ht]
    \centering
    \includegraphics[width=0.70\linewidth]{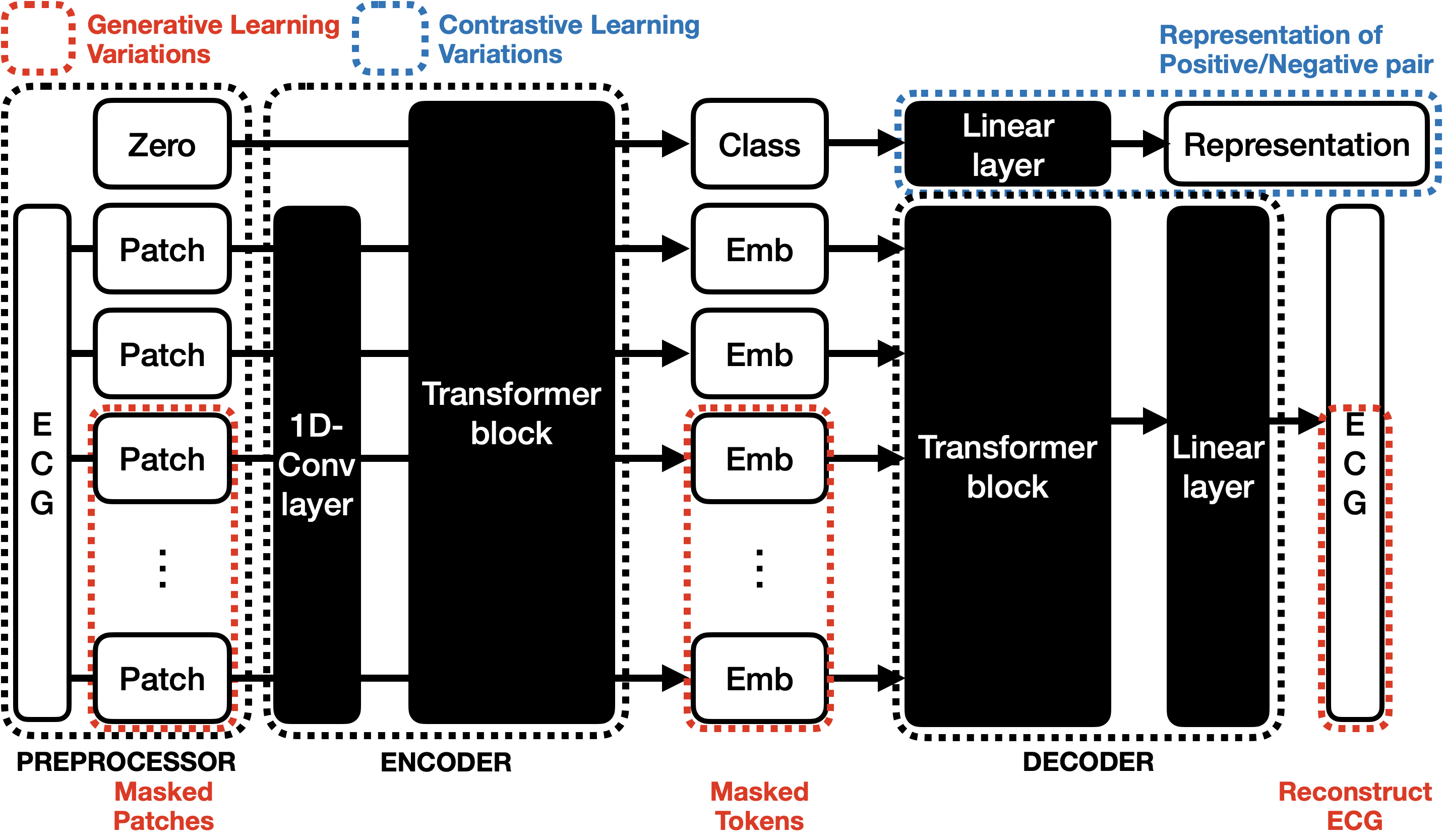}
    \caption{Structure of SiT comprises three key components: a shared encoder, a representor, and a decoder. Training paths for GL and CL are represented by red and blue arrows, respectively.}
    \label{fig:overview}
    \vspace{-0.2cm}
\end{figure*}

\section{Contrastive Regularized Masked Autoencoder} 
In this study, we propose \textbf{CREMA} (\textbf{C}ontrastive \textbf{Re}gularized \textbf{M}asked \textbf{A}utoencoder), a pre-trained model that integrates generative learning (GL) and contrastive learning (CL) to learn generalizable ECG representations. CREMA builds on the \textbf{\textit{SiT}} architecture, which extends the Vision Transformer (ViT)~\cite{dosovitskiy2020image} with a one-dimensional convolution block to better encode the morphology of ECG signals. As illustrated in Figure~\ref{fig:overview}, the SiT consists of three components: a shared encoder, a representor, and a decoder. The GL and CL training paths are depicted in red and blue arrows, respectively. In this section, we highlight the key architectural difference from the original ViT—namely, the shared encoder that combines convolution and transformer modules.

\subsection{Shared Encoder}
We designed the shared encoder to capture both local and global features of ECG signals by combining a 1D convolution block and a transformer block. The 1D convolution block provides an \textbf{inductive bias toward local pattern recognition}—including translation invariance and weight sharing—which improves data efficiency and stability in structured signals~\cite{dosovitskiy2020image, grover2023bias}. This bias is particularly effective for detecting localized waveform components such as P waves, QRS complexes, and T waves, as shown in recent ECG studies~\cite{song2023realtime, saha2025lightweight}. However, 1D convolutions are limited in modeling dependencies across cardiac cycles. To address this, the transformer block captures \textbf{sequence-level dynamics}, such as rhythm regularity and inter-beat intervals, using self-attention~\cite{vaswani2017attention}. A class token, initialized from a normal distribution, is prepended to the patch embeddings and passed through the transformer, producing both a global ECG representation and contextualized patch features. This architecture combines the efficiency of 1D convolutions with the expressive capacity of attention to encode the multi-scale nature of ECG signals.

\vspace{-0.2cm}
\subsection{Training Path}


\noindent\textbf{\textit{GL Path:}}  
The input ECG is used without augmentation and is divided into temporal patches designed to capture local morphological features. A standard 12-lead ECG typically includes at least one heartbeat per second, containing essential components such as P, QRS, and T waves. The patch size is chosen to cover at least one cardiac cycle segment (P–T) and, at most, one complete beat (e.g., a maximum patch size of 250 for an input length of 2500).

A random subset of patches is masked at a predefined ratio (e.g., 75\%). The shared encoder converts the visible patches into patch embeddings, which are then passed to the decoder for reconstruction. Only the patch embeddings are used in the decoding process, excluding the class token. The reconstruction loss is computed using the Mean Absolute Error (MAE) and is backpropagated to optimize the encoder and decoder, promoting the model's ability to learn high-fidelity morphological representations.

\vspace{0.2cm}


\noindent\textbf{\textit{CL Path:}}  
To facilitate contrastive learning, paired views of each input ECG are generated via two strategies: \textit{sample-level augmentation} and \textit{patient-level pairing}. At the sample level, augmentations such as Time Mask, Channel Mask, Baseline Wander, Baseline Shift, Partial White Noise, and EMGNoise~\cite{lee2022efficient} are randomly applied to create a perturbed view. At the patient level, another ECG recorded from the same individual at a different time is selected as the positive pair.

Both the original and augmented ECGs are patchified, masked (as in the GL path), and passed through the shared encoder to obtain class and patch embeddings. The class embeddings are refined by the representor to form discriminative representations. A contrastive loss (e.g., NT-Xent) is applied to align the paired embeddings while pushing apart negatives. This loss is backpropagated to optimize the encoder and representor toward learning representations with improved inter-class separability.

\subsection{Contrastive Regularized MAE Loss}
GL captures details of ECGs via reconstruction, but often yields dense embeddings with limited discriminative power~\cite{radhakrishnan2023cross}. CL enhances separability but may distort subtle ECG features due to augmentation~\cite{na2024guiding, liu2024zero}.
To balance these objectives, we propose the \textbf{Contrastive Regularized MAE loss}, which combines GL and CL to produce representations that are both precise and semantically structured.

The reconstruction loss, calculated as the mean absolute error (MAE), quantifies the discrepancy between the input ECG ($x_i$) and reconstructed ECG ($y_i$), which is defined as:
\begin{equation}
    \mathcal{L}_{R} = \sum_{i=1}^{N} |x_i - y_i|    
    \label{eq:recon_loss}
\end{equation}
where $N$ denotes the total number of ECG samples.

The contrastive loss quantifies the differences between similar and dissimilar paired views of ECG representations. It is calculated using the NT-Xent (Normalized Temperature-scaled Cross Entropy) loss function~\cite{chen2020simple} as:
\begin{equation}
    \mathcal{L}_{C} = -\log \frac{\exp(sim(z_i, z_j) / \tau)}{\sum_{k=1}^{2N} \mathbbm{1}_{[i\ne k]} \exp(sim(z_i, z_k) / \tau)}    
    \label{eq:cont_loss}
\end{equation}
Here, $z_i$ and $z_j$ denote the representations of the positive pair, $sim(z_i, z_j)$ does a cosine similarity of $z_i$ and $z_j$, $tau$ does a temperature parameter, and $N$ does the number of ECG samples.

To balance morphological fidelity of GL and semantic separability of CL, we define the \textbf{Contrastive Regularized MAE loss} as:
\begin{equation}
\mathcal{L}_{\text{CREMA}} = \mathcal{L}_{R} + \lambda \left( \mathcal{L}_{C_\text{sample}} + \mathcal{L}_{C_\text{patient}} \right),
\label{eq:crema_loss}
\end{equation}

where $\mathcal{L}_R$ encourages faithful reconstruction and $\mathcal{L}_C$ promotes discriminative structure through contrastive regularization. From an information-theoretic perspective, this loss can be interpreted as maximizing mutual information $I(z; x)$ under a separability constraint:
\begin{equation}
\max_{z} \quad I(z; x) \quad \text{s.t.} \quad D(z^+, z^-) \geq \delta,
\end{equation}

\begin{table*}[]
\small
\caption{Linear probing performance (AUROC) of CREMA and other SSL methods on PTB-XL and CPSC2018 datasets using 1\%, 10\%, and 100\% of labeled data. Random Init and CREMA w/o $\lambda$ are evaluated only at 100\% due to their roles as baseline and ablation references, respectively. Best results are in bold, second-best are underlined.}
\label{tab:perform_linear}
\begin{tabular}{@{}l|rrr|rrr|rrr|rrr|rrr@{}}
\toprule
& \multicolumn{3}{c|}{\textbf{PTB-XL Super}}& \multicolumn{3}{c|}{\textbf{PTB-XL Sub}}& \multicolumn{3}{c|}{\textbf{PTB-XL Form}}& \multicolumn{3}{c|}{\textbf{PTB-XL Rhythm}}& \multicolumn{3}{c}{\textbf{CPSC2018}}\\ \midrule
\textbf{Method (Backbone)} & \textbf{1\%} & \textbf{10\%} & \textbf{100\%} & \textbf{1\%} & \textbf{10\%} & \textbf{100\%} & \textbf{1\%} & \textbf{10\%} & \textbf{100\%} & \textbf{1\%} & \textbf{10\%} & \textbf{100\%} & \textbf{1\%} & \textbf{10\%} & \textbf{100\%} \\ \midrule
\textbf{Rand. Init. (SiT)} & - & - & 86.76 & - & - & 85.40 & - & - & 63.05 & - & - & 86.08 & - & - & 74.30 \\ \midrule
\textbf{SimCLR (SiT)} & 82.71 & 86.77 & 89.24 & 66.65 & 75.67 & 87.13 & 54.48 & \underline{68.56} & 74.75 & 65.94 & 72.21 & 83.07 & \underline{70.78} & 80.00 & 87.03 \\
\textbf{CLOCs (SiT)} & 82.01 & 87.47 & 89.72 & \underline{70.10} & \underline{79.72} & 90.43 & 57.20 & 68.22 & 77.49 & 71.02 & 77.63 & 86.34 & 68.18 & \underline{82.93} & 90.24 \\
\textbf{COMET (SiT)} & 82.34 & \underline{88.69} & 90.44 & \textbf{71.09} & \textbf{80.51} & \underline{90.57} & \underline{57.55} & 62.59 & 75.01 & 69.17 & 76.41 & 86.00 &  69.35 & 82.30 & 90.28 \\ 
\textbf{ST-MEM (ViT)} & 69.14 & 71.42 & 81.49 & 66.76 & 71.18 & 78.52 & 57.52 & 62.02 & 66.25 & 66.07 & 71.03 & 82.53 & 66.22 & 71.75 & 79.05 \\ \midrule
\textbf{CMAE (SiT)} & \underline{82.42} & 85.73 & 87.55 & 64.95 & 74.76 & 88.10 & 55.02 & 64.95 & 77.39 & \textbf{72.20} & \underline{80.11} & \underline{87.07} & 68.95 & 79.32 & 87.72 \\ 
\textbf{CREMA w/o $\lambda$ (SiT)} & - & - & \underline{90.52} & - & - & 90.16 & - & - & \underline{78.79} & - & - & 80.57 & - & - & \underline{91.01}\\ 
\textbf{CREMA (SiT)} & \textbf{83.98} & \textbf{88.97} & \textbf{91.25} & 65.49 & 78.04 & \textbf{91.37} & \textbf{60.71} & \textbf{69.79} & \textbf{80.14} & \underline{71.64} & \textbf{82.19} & \textbf{88.92} & \textbf{72.52} & \textbf{87.70} & \textbf{92.78} \\ \bottomrule
\end{tabular}
\vspace{-0.2cm}
\end{table*}

With $\lambda$ acting as a Lagrangian multiplier. Low values of $\lambda$ may lead to over-preserved, dense embeddings; high values may impair reconstruction. A well-chosen $\lambda$ yields representations that are both precise and discriminative, enabling effective modeling of clinically meaningful ECG variations.

\section{Evaluation}\label{sec:exp_result}
This section provides a summary of the evaluations and results. For performance metrics, we utilized the average of Area Under the Receiver Operating Characteristic Curve (AUROC) on multi-label classifications across downstream tasks. The average AUROC indicates \textit{\textbf{importance of ensuring that a pre-trained model demonstrates consistent performance across all}} rather than excelling in a single task.

\subsection{Dataset} 
\noindent\textit{\textbf{Pre-training dataset:}} We used datasets compiled from five public repositories: MIMIC~\cite{johnson2016mimic}, CODE15~\cite{ribeiro_2021_4916206}, BIOBANK~\cite{sudlow2015uk}, SAMI~\cite{ribeiro_2021_4905618}, and IKEM~\cite{michal_sejak_2023_8393007}. These datasets were chosen to account for \textbf{demographic diversity} by including data collected from multiple continents. The combined dataset comprises 1,291,868 ECG samples from 442,736 distinct patients, reducing potential biases and enhancing the generalizability. 

\vspace{0.1cm}
\noindent\textit{\textbf{Downstream dataset:}} We used datasets from two public repositories: PTB-XL~\cite{wagner2020ptb} and CPSC2018~\cite{liu2018open}. 
The PTB-XL dataset includes a total of 21,837 ECG samples from 18,885 patients and four subsets for multi-label classification: \textbf{\textit{Superclass}} (5 labels), \textbf{\textit{Subclass}} (23 labels), \textbf{\textit{Form}} (19 labels), and \textbf{\textit{Rhythm}} (12 labels). We follow the official data split for training, validation, and testing~\cite{wagner2020ptb}.
The CPSC2018 dataset includes 6,877 ECG samples and \textbf{\textit{nine distinct labels}}. We follow the prior settings~\cite{liu2018open} for data split, which randomly splits into 7:1:2 for training, validation, and testing. 

Before the experiment, all data were standardized to ensure consistency in sample rate and measurement duration. The sample rate was set to 250 Hz, with 10 seconds resulting in ECG signals with 2,500 data points per lead. Additional details are provided in Table~\ref{tab:summary_dataset} in Appendix~\ref{apdx:dataset}.

\subsection{Implementation}\label{sec:implementation}
To evaluate CREMA, we compare it against established SSL methods with complementary designs: ST-MEM~\cite{na2024guiding}, MAE~\cite{he2022masked}, SimCLR~\cite{chen2020simple}, CLoCS~\cite{kiyasseh2020clocs}, and COMET~\cite{wang2023contrast}. To ensure a fair comparison, we use each method's original augmentation strategy and backbone: ST-MEM uses ViT-B~\cite{na2024guiding}, while SimCLR, CLoCS, COMET, CMAE, and CREMA are implemented with our unified SiT backbone (Figure~\ref{fig:overview}).

Augmentation policies follow the original works: SimCLR, CLoCS, and COMET apply Cutout, Drop, and Gaussian Noise, while CREMA uses Time Mask, Channel Mask, Baseline Wander, Baseline Shift, Partial White Noise, and EMGNoise~\cite{lee2022efficient}. COMET's trial-level contrastive objective was omitted due to the lack of trial metadata in our unlabeled set. MAE, originally designed for vision tasks, is adapted to ECGs as CMAE.

For downstream evaluation, each pre-trained model is assessed using both linear probing and fine-tuning. In linear probing, the encoder is frozen and only a linear classifier is trained to measure representation quality. In fine-tuning, all weights, including the encoder and classifier, are updated to adapt fully to the task.

\begin{table*}[]
\small
\caption{Fine-tuning performance (AUROC) of supervised and pre-trained models on PTB-XL and CPSC2018. CREMA w/o $\lambda$ is an ablation variant without contrastive regularization. Best results are in bold, second-best are underlined.}
\label{tab:perform_fine}
\begin{tabular}{@{}l|r|r|r|r|r@{}}
\toprule
\textbf{Method (Backbone)} & \textbf{PTB-XL Super} & \textbf{PTB-XL Sub} & \textbf{PTB-XL Form} & \textbf{PTB-XL Rhythm} & \textbf{CPSC2018} \\ \midrule
\textbf{Scratch (SiT)} & 91.78 & 90.84 & 81.77 & 92.14 & 93.82 \\ 
\textbf{Scratch (ViT)} & 86.98 & 85.07 & 75.49 & 90.30 & 91.37 \\ \midrule
\textbf{SimCLR (SiT)} & 92.28 & 92.04 & 84.66 & \underline{92.63} & 94.33 \\
\textbf{CLOCs (SiT)}  & 92.09 & 90.78 & 80.43 & 91.14 & 93.87 \\
\textbf{COMET (SiT)}  & 92.30 & 92.20 & 78.98 & 91.88 & 94.21 \\ 
\textbf{ST-MEM (ViT)} & 87.95 & 87.84 & 72.98 & 90.68 & 93.20 \\ \midrule
\textbf{CMAE (SiT)}   & 92.23 & 91.12 & \underline{85.81} & 91.29 & 95.04 \\ 
\textbf{CREMA w/o $\lambda$ (SiT)} & \underline{92.30} & \underline{92.95} & 83.15 & 91.94 & \underline{95.67} \\ 
\textbf{CREMA (SiT)}  & \textbf{92.86} & \textbf{93.35} & \textbf{87.07} & \textbf{93.13} & \textbf{95.77} \\ \bottomrule
\end{tabular}
\vspace{-0.2cm}
\end{table*}

\subsection{Linear Probing Evaluation}\label{sec:linear_eval}
Table~\ref{tab:perform_linear} compares the linear probing performance of CREMA and other pre-trained models against a baseline model with a randomly initialized SiT encoder. All pre-trained models, except ST-MEM, outperform the baseline across all downstream tasks. ST-MEM’s relatively lower performance is likely due to differences in backbone architecture, while CMAE, which uses the same SiT backbone, consistently surpasses the baseline. Notably, CMAE achieved the second-best performance after CREMA in the rhythm classification task on PTB-XL, consistent with prior findings.

With 100\% labeled data, CREMA achieves the highest performance across all tasks—including diagnostic superclass and subclass classification, morphological form, and rhythm pattern detection—and consistently outperforms all models on the diverse CPSC2018 dataset. However, under limited data conditions (1\% and 10\%), CREMA shows lower performance in subclass classification, where the number of labels is most significant (i.e., 23 labels). In these settings, CLoCS and COMET perform better, suggesting that contrastive strategies are particularly effective at enhancing discriminative capacity when labels are scarce.

CREMA’s performance is further improved when contrastive and generative losses are properly balanced using the $\lambda$ parameter, underscoring the importance of the proposed contrastive regularized MAE loss.

Overall, these results demonstrate that CREMA not only achieves strong and consistent performance across diverse ECG classification tasks but also highlights the effectiveness of the SiT backbone in capturing both local and global ECG features, supporting the generalizability of the learned representations.

\subsection{Fine-tuning Evaluation}\label{sec:finetune_eval}
Table~\ref{tab:perform_fine} presents the performance comparison between supervised models trained from scratch and fine-tuned pre-trained models. The SiT-based scratch model demonstrates competitive performance and, in specific tasks, slightly outperforms some pre-trained methods. Notably, ST-MEM underperforms, while CMAE consistently exceeds the baseline, likely due to architectural differences, as also observed in linear probing.

The scratch results also highlight the representational strength of the SiT backbone. Compared to ViT under identical training conditions, the SiT-based model achieves higher performance across all tasks on PTB-XL and CPSC2018. This gap illustrates the structural advantages of SiT, particularly its ability to extract both local waveform details and global rhythm patterns via integrated 1D convolution. Even without pretraining, SiT effectively encodes clinically relevant ECG features, reinforcing its suitability as a backbone for ECG modeling.

Among pre-trained models, CREMA achieves the best results across all tasks. Ablation shows that removing the $\lambda$ parameter, applying equal weighting (1:1) to contrastive and generative losses, leads to consistent performance degradation, confirming the importance of balanced objective design. Nonetheless, CREMA without $\lambda$ still outperforms CMAE, which uses only generative learning, highlighting the benefit of incorporating contrastive regularization.

In summary, CREMA’s strong performance stems from its balanced learning objective, integrating contrastive regularization with generative reconstruction, and its SiT backbone, which jointly supports robust and generalizable ECG representation learning.

\subsection{Robustness on Distribution Shift}
To evaluate the robustness of the learned ECG representation across different sources, we conduct linear probing with
SSL methods and CREMA under domain shifts: training on one dataset (i.e., source domain) and testing
on another (i.e., target domain), with categories in common with the source domain.

\begin{table}[!h]
\vspace{-0.1cm}
\small
\caption{Performance (AUROC) under distribution shift. ‘Source’ indicates the dataset used for linear probing; ‘Target’ is the corresponding test set with matched categories. Best results are in bold, second-best are underlined.}
\label{tab:domain_shift}
\begin{tabular}{@{}l|r|r@{}}
\toprule
\textbf{Source domain} & \textbf{CPSC2018} & \textbf{PTB-XL Supe}r    \\ \midrule
\textbf{Target domain} & \textbf{PTB-XL Super}  & \textbf{CPSC2018} \\ \midrule
\textbf{SimCLR} & 60.31 & 81.90 \\
\textbf{CLOCs}  & \underline{63.26} & 82.34 \\
\textbf{CMAE}   & 57.64 & \underline{82.65} \\
\textbf{COMET}  & 61.66 & 81.59 \\ 
\textbf{ST-MEM} & 62.27 & 76.12 \\ \midrule
\textbf{CREMA}  & \textbf{65.68} & \textbf{84.27} \\ \bottomrule
\end{tabular}
\vspace{-0.2cm}
\end{table}

We follow the target domain preparation protocol~\cite{liu2024zero}. After preparing the target domain samples, we compare CREMA with all SSL methods using 100\% data for linear probing across target domains. The results are summarized in Table~\ref{tab:domain_shift}. Remarkably, CREMA outperforms all SSL methods in linear probing evaluation. 

We also confirm that CLOCs achieves the second-highest on CPSC2018 to PTB-XL Super and CMAE on PTB-XL Super to CPSC2018. This may be because PTB-XL Super has diagnostic labels that are advantageous for discrimination tasks in contrastive learning, while CPSC2018 has morphology, rhythm, and diagnostic labels that are advantageous for reconstruction tasks in generative learning.

These results suggest that either only contrastive learning or only generative learning may hinder the robustness of the ECG representation~\cite{na2024guiding, radhakrishnan2023cross}. On the other hand, the results of CRMEA, which uses both learning methods in a balanced manner, show that the learned ECG features are both representative and robust.

\subsection{Advantages of Contrastive Regularization}~\label{sec:advantage_loss}

This section analyzes the effect of the trade-off parameter $\lambda$ in the contrastive-regularized MAE loss (Equation~\ref{eq:crema_loss}). We varied $\lambda$ from 0 to 2 and evaluated validation loss trends across its components after 50 training epochs. When $\lambda=0$, only the generative loss is used—equivalent to CMAE (Section~\ref{sec:implementation}). As $\lambda$ increases beyond 1, the contrastive losses receive greater relative weighting, shifting the objective toward representation separability. All loss values were min-max normalized for comparability.

\begin{figure}[!h]
    \vspace{-0.1cm}
    \centering
    \includegraphics[width=.8\linewidth]{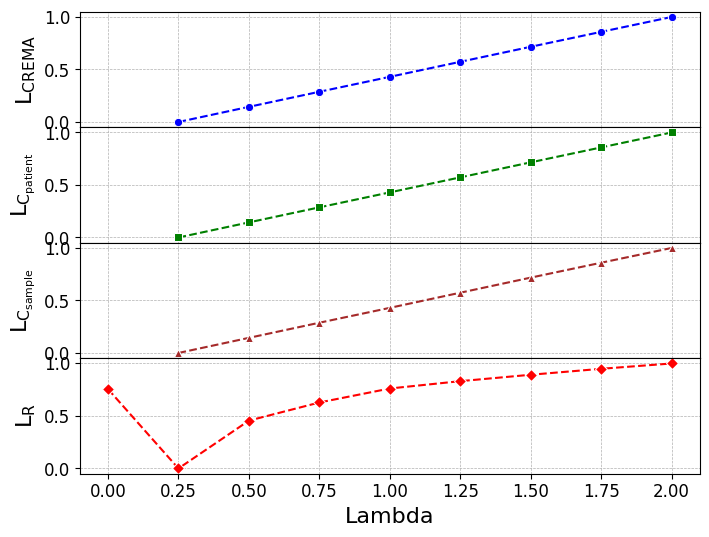}
    \caption{The change of the losses on the validation set after 50 training epochs, according to the varying lambda 0 to 2.}
    \label{fig:loss_lambda}
    \vspace{-0.2cm}
\end{figure}

Figure~\ref{fig:loss_lambda} plots each loss component as a function of $\lambda$: $L_{\text{CREMA}}$, $L_{C_{\text{patient}}}$, $L_{C_{\text{sample}}}$, and $L_R$. The total loss $L_{\text{CREMA}}$ is minimized at $\lambda=0.25$, where $L_R$ is also lower than at $\lambda=0$, indicating \textbf{\textit{reduced overfitting to reconstruction}}. As $\lambda$ increases, $L_{C_{\text{patient}}}$ and $L_{C_{\text{sample}}}$ rise approximately linearly, while $L_R$ decreases initially but grows rapidly beyond $\lambda=0.25$.

This suggests that $\lambda=0.25$ offers a practical balance between preserving ECG morphology and enhancing representation separability. Smaller values result in dense and less informative embeddings, while larger values compromise reconstruction quality. The trade-off controlled by $\lambda$ determines how the model prioritizes between generative fidelity and contrastive discrimination.

\begin{table}[!h]
\small
\caption{Ablation results showing the contribution of sample-level contrast, patient-level contrast, and weighted regularization ($\lambda$) to overall CREMA performance (AUROC).}
\label{tab:ablation}
\begin{tabular}{@{}l|rr@{}}
\toprule
 & \textbf{PTB-XL Super} & \textbf{CPSC2018} \\ \midrule
\textbf{GL (CMAE)} & 92.23 & 95.04 \\
\textbf{\& Sample-level CL} & 92.28 & 95.59 \\
\textbf{\& Patient-level CL} & 92.30 & 95.67 \\
\textbf{\& $\lambda$ (CREMA)} & 92.86 & 95.77 \\ \bottomrule
\end{tabular}
\vspace{-0.2cm}
\end{table}

This trend is reinforced by the ablation study in Table~\ref{tab:ablation}. Starting from the generative-only baseline (CMAE: 92.23 on PTB-XL, 95.04 on CPSC2018), adding sample-level contrastive learning yields incremental improvements (92.28/95.59), with further gains from incorporating patient-level contrast (92.30/95.67). The full model with the proposed weighting scheme ($\lambda=0.25$) achieves the best results (92.86/95.77), outperforming all intermediate variants. This progression highlights that the weighting mechanism, rather than mere inclusion of contrastive signals, is key to maximizing performance.

Overall, the results demonstrate that contrastive regularization, when selectively structured and properly weighted, improves generalizability by aligning semantic separability with morphological preservation. Excessive contrastive emphasis ($\lambda > 1$), however, substantially degrades reconstruction, underscoring the importance of balanced objective design.

\section{Deployment}

\begin{figure}[!h]
    \vspace{-0.2cm}
    \centering
    \includegraphics[width=.7\linewidth]{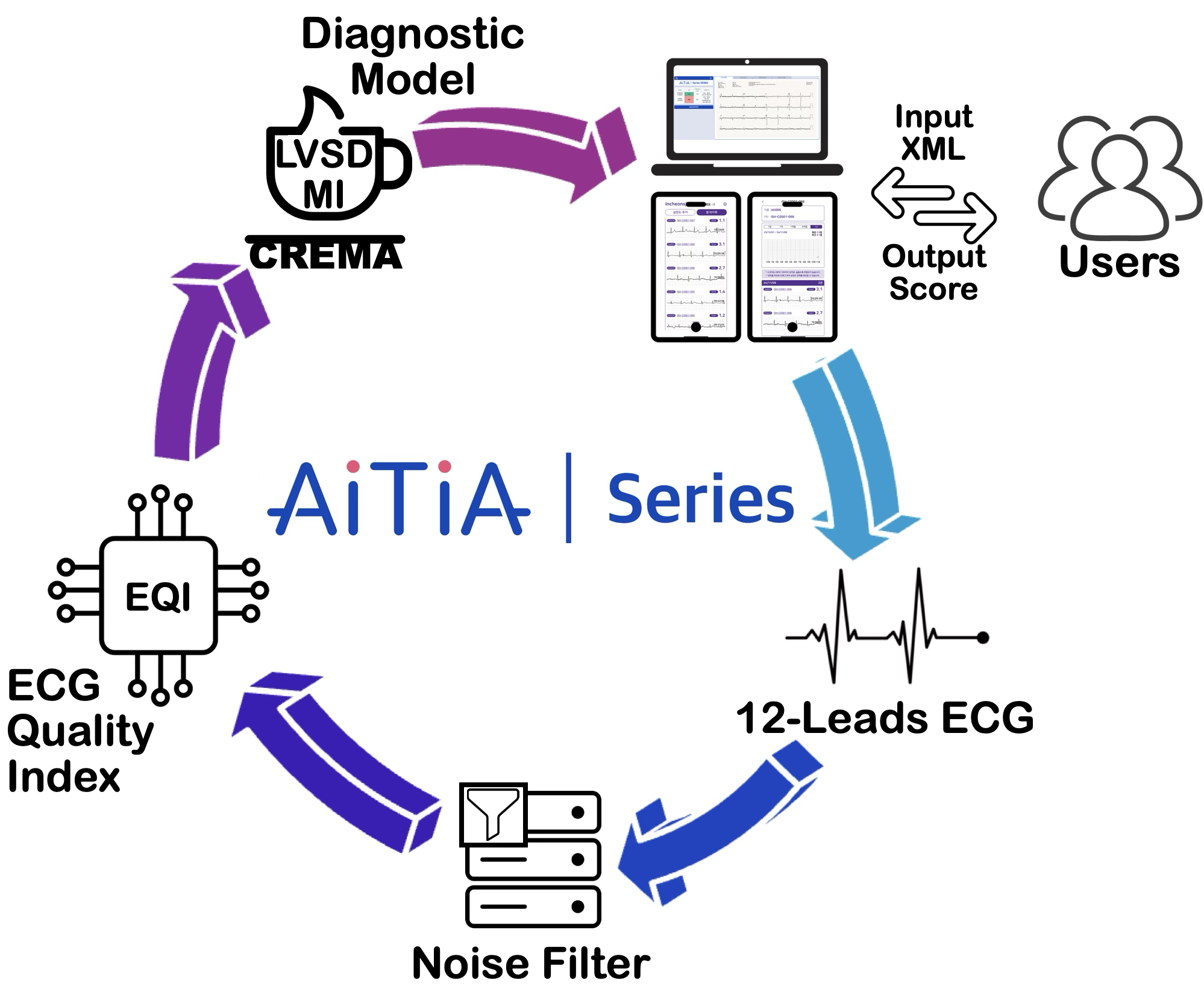}
    \caption{Overview of AiTiA-Series. Click the next URL to open the demo page: \url{https://aitia-demo.medicalai.com}.} 
    \label{fig:deployment}
    \vspace{-0.2cm}
\end{figure}

Figure~\ref{fig:deployment} provides an overview of AiTiA-Series, a cardiac disease diagnosis assistance service utilizing standard 12-lead ECG data. Currently deployed in over 50 medical institutions, including major hospitals across South Korea, the AiTiA-Series features a comprehensive suite of tools, including web and mobile application interfaces, a noise filtering system, an ECG Quality Index (EQI) model, and an advanced diagnostic models: LVSD and MI, fine-tuning CREMA as the foundation model.

Users upload ECGs recorded by electrocardiographs in XML format via the interface, which can also integrate automatically with ECG devices such as 12-lead electrocardiographs. The system applies noise filtering and assesses the data quality using the ECG Quality Index (EQI). If the ECG quality meets the threshold, the diagnostic model determines whether the target cardiac disease is present and delivers the result to the user through the interface.

\begin{table*}[!h]
\vspace{-0.2cm}
\small
\caption{Overview of the clinical dataset for LVSD/MI diagnosis in AiTiA-Series; source identifiers use numbers to represent distinct medical institutions, and “GC” and “ER” indicate general clinic and emergency room departments, respectively.}
\label{tab:clinical_data}
\begin{tabular}{@{}l|rrrc|rrrc@{}}
\toprule
\multirow{2}{*}{} & \multicolumn{4}{c|}{\textbf{LVSD}} & \multicolumn{4}{c}{\textbf{MI}} \\ \cmidrule(l){2-9} 
 & \multicolumn{1}{c}{\textbf{\# Sample}} & \multicolumn{1}{c}{\textbf{\# Patients}} & \multicolumn{1}{c}{\textbf{\# Case (Ratio)}} & \multicolumn{1}{c|}{\textbf{Source ID}} & \multicolumn{1}{c}{\textbf{\# Sample}} & \multicolumn{1}{c}{\textbf{\# Patients}} & \multicolumn{1}{c}{\textbf{\# Case (Ratio)}} & \multicolumn{1}{c}{\textbf{Source ID}}\\ \midrule
\textbf{Train} & 400,339 & 148,624 & 49,757 (12\%) & \multirow{3}{*}{GC.0/1/2/3} & 36,170 & 24,824 & 11,327 (31.3\%) & \multirow{3}{*}{GC.0/1/2/3} \\
\textbf{Validation} & 49,247 & 19,054 & 5,573 (11\%) & & 4,019 & 3,772 & 1,259 (31.3\%) & \\
\textbf{Internal Test} & 49,140 & 19,211 & 5,798 (12\%) & & 4,119 & 2,870 & 1,153 (28\%) & \\ \midrule
\textbf{External Test} & 79,605 & 42,709 & 9,261 (12\%) & \multirow{1}{*}{GC.4/5, ER.6/7} & 3,363 & 2,193 & 599 (17\%) & \multirow{1}{*}{GC.4} \\ \bottomrule
\end{tabular}
\vspace{-0.2cm}
\end{table*}

\subsection{Clinical Dataset}

The diagnostic models were fine-tuned using a clinical dataset comprising 498,726 samples for LVSD and 44,308 for MI, collected from multiple hospitals and clinics across South Korea. In the source identifiers, numeric codes represent distinct medical institutions, while “GC” and “ER” indicate general clinic and emergency room departments, respectively. Details are provided in Table~\ref{tab:clinical_data}.

The dataset reflects the typical imbalance found in medical data, with positive case ratios of 12\% for LVSD and 31\% for MI in the training set. Data were systematically partitioned into training, validation, and internal test sets, each sourced from general patient populations across distinct medical institutions.

In addition, an external test set was collected from live-served clinical settings, including 79,605 samples for LVSD (12\%) and 3,363 for MI (18\%). This set includes data from emergency departments (ER.6 and ER.7), which differ in patient demographics and clinical context from the internal data. These distributional differences emphasize the need to assess model robustness under realistic deployment conditions.

The classifier architecture described in Section~\ref{sec:implementation} was applied consistently across all fine-tunings.

\subsection{Performance on Clinical Environments}\label{sec:clinical_eval}

Table~\ref{tab:deployment_exp} presents a performance comparison between the supervised baseline (1D-ResNet50) and the proposed CREMA model for LVSD and MI diagnosis, evaluated on both internal and external datasets. The internal set follows the same distribution as the training data, while the external set is sourced from real-world clinical environments, introducing a meaningful distribution shift.

\begin{table}[!h]
\vspace{-0.1cm}
\small
\caption{Performance of AiTiA-LVSD/MI on internal (source) and external (target) datasets. The internal set shares the training distribution, while the external set reflects live-served clinical deployment.}
\label{tab:deployment_exp}
\begin{tabular}{@{}cc|rrrr@{}}
\toprule
\multirow{2}{*}{\textbf{\begin{tabular}[c]{@{}c@{}}Cardiac\\ Disease\end{tabular}}} &
  \multirow{2}{*}{\textbf{Model}} &
  \multicolumn{2}{c}{\textbf{Internal}} &
  \multicolumn{2}{c}{\textbf{External}} \\ \cmidrule(l){3-6} 
 &
   &
  \multicolumn{1}{c}{\textbf{AUROC}} &
  \multicolumn{1}{c}{\textbf{AUPRC}} &
  \multicolumn{1}{c}{\textbf{AUROC}} &
  \multicolumn{1}{c}{\textbf{AUPRC}} \\ \midrule
\multirow{3}{*}{\textbf{LVSD}} & \textbf{1D-ResNet50} & 0.938   & 0.712   & 0.947   & 0.758   \\
                               & \textbf{CREMA}       & 0.947   & 0.743   & 0.960   & 0.804   \\ \cmidrule{2-6}
                               & \textbf{Gain}        & +0.85\% & +4.06\% & +1.37\% & +6.07\% \\ \midrule
\multirow{3}{*}{\textbf{MI}}   & \textbf{1D-ResNet50} & 0.946   & 0.873   & 0.956   & 0.886   \\
                               & \textbf{CREMA}       & 0.953   & 0.903   & 0.965   & 0.919   \\ \cmidrule{2-6}
                               & \textbf{Gain}        & +0.74\% & +3.44\% & +0.94\% & +3.72\% \\ \bottomrule
\end{tabular}
\vspace{-0.2cm}
\end{table}

Across both tasks, CREMA consistently outperforms the baseline regarding AUROC and AUPRC. While AUROC improvements are modest (+0.74\% to +1.37\%), AUPRC gains are substantially larger—up to +6.07\% for LVSD and +3.72\% for MI on the external set. This disparity highlights CREMA’s enhanced ability to detect true positive cases, an advantage in class-imbalanced clinical data.

The marked improvements in AUPRC demonstrate that CREMA achieves higher precision and recall for the minority (disease-positive) class, even under domain shift. In particular, the 6.07\% AUPRC increase for LVSD on the external set suggests CREMA’s suitability for deployment in critical clinical scenarios such as emergency care, where diagnostic accuracy is vital.

These results indicate that while AUROC reflects overall discrimination, AUPRC better captures real-world clinical utility. The consistent improvements reaffirm that CREMA learns robust and generalizable ECG representations that remain effective across diverse data distributions, enhancing diagnostic reliability in deployment settings.

Table~\ref{tab:detail_lvsd} further breaks down CREMA’s performance for LVSD across four distinct medical institutions (4, 5, 6, and 7) in the external test set. Among these, GC.4 and GC.5 represent the general department, while ER.6 and ER.7 correspond to emergency departments, which typically involve higher-acuity cases and distinct clinical conditions.

\begin{table}[!h]
\vspace{-0.1cm}
\small
\caption{LVSD performance on the external test set; source number denotes distinct medical inst., and “GC” and “ER” indicate general and emergency departments, respectively.}
\label{tab:detail_lvsd}
\begin{tabular}{@{}c|c|rr@{}}
\toprule
\multirow{2}{*}{\textbf{}} & \multirow{2}{*}{\textbf{\begin{tabular}[c]{@{}c@{}}Source\\ ID.\end{tabular}}} & \multicolumn{2}{c}{\textbf{LVSD}} \\ \cmidrule(l){3-4} 
 & & \multicolumn{1}{c}{\textbf{AUROC}} & \multicolumn{1}{c}{\textbf{AUPRC}} \\ \midrule
\multirow{4}{*}{\textbf{CREMA}} & \textbf{GC.4} & 0.962 & 0.805 \\
 & \textbf{GC.5} & 0.942 & 0.772 \\
 & \textbf{ER.6} & 0.939 & 0.839 \\
 & \textbf{ER.7} & 0.952 & 0.840 \\ \bottomrule
\end{tabular}
\vspace{-0.3cm}
\end{table}

Although CREMA was pre-trained and fine-tuned solely on data from general institutions, it performs consistently well across all settings, including emergency departments. AUROC remains high across the board (0.939–0.962), while AUPRC is notably higher in ER settings (0.839 and 0.840) compared to general institutions (0.805 and 0.772). This suggests that CREMA is particularly effective at identifying LVSD cases in clinically complex environments. 

Overall, these findings underscore CREMA’s robustness to distribution shift and its strong generalizability from training domains to deployment contexts that differ in patient characteristics and disease manifestation. Its stable performance in emergency settings, despite training only on general data, demonstrates its practical utility and reliability for real-world clinical applications.

\section{Conclusion}
This study introduced CREMA (Contrastive Regularized Masked Autoencoder), a self-supervised foundation model for standard 12-lead ECGs. Built on a straightforward yet expressive SiT architecture, CREMA captures both local morphology and global rhythm by integrating contrastive and generative learning.
CREMA demonstrates strong generalizability, outperforming supervised and existing SSL methods across multiple ECG classification tasks. It achieves notable gains in linear probing and fine-tuning, particularly under low-label settings and distribution shifts, and shows reliable performance in real-world clinical deployment. These results highlight CREMA’s scalability, efficiency, and practicality as a foundation model for ECG diagnostics.
While currently focused on disease classification, future work may extend CREMA to broader ECG tasks and improve interpretability for clinical trust. Our findings establish CREMA as a new benchmark for generalizable and deployable ECG representation learning.

\newpage

\bibliographystyle{ACM-Reference-Format}
\bibliography{sample-base}

\newpage
\appendix

\renewcommand{\thesection}{\Alph{section}}
\setcounter{section}{0}

\section{Supplemental Results}\label{sec:supplemental_result}

\subsection{Risk Decomposition Analysis}
This section analyzes the potential limitations of foundation models for ECG diagnostics through a quantitative lens by decomposing total predictive error into interpretable components. We apply risk decomposition~\cite{dubois2023evaluating} to the linear probing models across four ECG classification tasks (MI, STTC, CD, and HYP), providing a detailed breakdown of performance across distinct sources of error.

The total error is divided into four components:
\begin{enumerate}
    \item \textbf{Approximation Error}, indicating model capacity to approximate the task function;
    \item \textbf{Representation Usability Error}, reflecting the suitability of the learned representations for downstream classification;
    \item \textbf{Probe Generalization Error}, capturing how well the linear classifier generalizes to unseen data;
    \item \textbf{Encoder Generalization Error}, quantifying the encoder’s robustness under distribution shifts.
\end{enumerate}

\begin{table}[!h]
\scriptsize
\centering
\caption{Results of risk decomposition applied to linear probing models for downstream ECG tasks on PTB-XL Super (MI, STTC, CD, HYP).}
\label{tab:risk_decomp}
\begin{tabular}{cc|rrrr|r}
\toprule
\multicolumn{2}{c}{} &
  \multicolumn{1}{|c}{\textbf{Approx.}} &
  \multicolumn{1}{c}{\begin{tabular}[c]{@{}l@{}}\textbf{Represent.}\\ \textbf{Usability.}\end{tabular}} &
  \multicolumn{1}{c}{\begin{tabular}[c]{@{}l@{}}\textbf{Probe.}\\ \textbf{Generaliz.}\end{tabular}} &
  \multicolumn{1}{c}{\begin{tabular}[c]{@{}l@{}}\textbf{Encoder.}\\ \textbf{Generaliz.}\end{tabular}} &
  \multicolumn{1}{|c}{\begin{tabular}[c]{@{}l@{}}\textbf{Total Risk.}\\ \textbf{(Total Error)}\end{tabular}} \\ \midrule
\multirow{5}{*}{\begin{tabular}[c]{@{}c@{}}\textbf{SimCLR}\\\textbf{(CL)}\end{tabular}} 
                    & \textbf{MI}   & 0.0930 & 0.1389 & 0.0080 & 0.0380 & 0.2779 \\
                    & \textbf{STTC} & 0.1159 & 0.0970 & 0.0870 & -0.0920 & 0.2080 \\
                    & \textbf{CD}   & 0.1110 & 0.0999 & -0.0309 & 0.0503 & 0.2303 \\
                    & \textbf{HYP}  & 0.1500 & 0.0669 & 0.0030 & 0.0106 & 0.2306 \\ \cmidrule{2-7}
                    & \textbf{AVR.}  & 0.1175 & \textbf{0.1007} & 0.0168 & 0.0017 & 0.2367 \\ \midrule
\multirow{5}{*}{\begin{tabular}[c]{@{}c@{}}\textbf{CMAE}\\\textbf{(GL)}\end{tabular}} 
                    & \textbf{MI}   & 0.1000 & 0.2289 & 0.0310 & -0.0010 & 0.3589 \\
                    & \textbf{STTC} & 0.1010 & 0.2009 & -0.0019 & -0.0190 & 0.2809 \\
                    & \textbf{CD}   & 0.1070 & 0.2560 & 0.0169 & -0.0517 & 0.3282 \\
                    & \textbf{HYP}  & 0.1459 & 0.1579 & 0.0160 & -0.0539 & 0.2660 \\ \cmidrule{2-7}
                    & \textbf{AVR.}  & \textbf{0.1135} & 0.2109 & \textbf{0.0155} & \textbf{-0.0314} & 0.3085 \\ \midrule
\multirow{5}{*}{\begin{tabular}[c]{@{}c@{}}\textbf{CREMA}\\\textbf{(HL)}\end{tabular}} 
                    & \textbf{MI}   & 0.0960 & 0.1509 & 0.0530 & -0.0430 & 0.2569 \\
                    & \textbf{STTC} & 0.1150 & 0.0909 & -0.0059 & 0.0209 & 0.2210 \\
                    & \textbf{CD}   & 0.1000 & 0.0819 & 0.0180 & -0.0124 & 0.1875 \\
                    & \textbf{HYP}  & 0.1480 & 0.1060 & 0.0659 & -0.0696 & 0.2503 \\ \cmidrule{2-7}
                    & \textbf{AVR.}  & 0.1148 & 0.1074 & 0.0328 & -0.0260 & \textbf{0.2289} \\ \bottomrule
\end{tabular}
\end{table}

Table~\ref{tab:risk_decomp} summarizes these components for SimCLR (contrastive), CMAE (generative), and CREMA (combined). SimCLR demonstrates relatively low total error and strong probe generalization, which aligns with its solid performance under low-label settings (see Section~\ref{sec:linear_eval}). CMAE, in contrast, shows high representation usability but suffers from weak generalization, especially in encoder-level robustness.

\begin{table*}[!h]
\small
\caption{Details of datasets used for pre-training and downstream task. \# invalid indicates the number of samples that are not included in any category.}
\label{tab:summary_dataset}
\begin{tabular}{@{}cccrrrrrrr@{}}
\toprule
 & \textbf{Name} & \textbf{Contry} & \multicolumn{1}{c}{\textbf{\# Patient}} & \multicolumn{1}{c}{\textbf{\# Label}} & \multicolumn{1}{c}{\textbf{\# Invalid}} & \multicolumn{1}{c}{\textbf{\# Train}} & \multicolumn{1}{c}{\textbf{\# Valid}} & \multicolumn{1}{c}{\textbf{\# Test}} & \multicolumn{1}{c}{\# \textbf{Total}} \\ \midrule
\multirow{5}{*}{\begin{tabular}[c]{@{}c@{}}\textbf{Pre-trained}\\ \textbf{Dataset}\end{tabular}} & \textbf{MIMIC} & \textbf{USA} & 161,352 & - & - & - & - & - & 800,035 \\
 & \textbf{CODE15} & \textbf{USA} & 233,770 & - & - & - & - & - & 341,292 \\
 & \textbf{BIOBANK} & \textbf{UK} & 15,365 & - & - & - & - & - & 50,780 \\
 & \textbf{SAMI} & \textbf{Brazil} & 1,959 & - & - & - & - & - & 1,631 \\
 & \textbf{IKEM} & \textbf{\begin{tabular}[c]{@{}c@{}}Czech\\ Repblic\end{tabular}} & 30,290 & - & - & - & - & - & 98,130 \\ \midrule
\multirow{5}{*}{\begin{tabular}[c]{@{}c@{}}\textbf{Downstream}\\ \textbf{Dataset}\end{tabular}} & \textbf{PTB-XL Super} & \textbf{Europe} & \multirow{4}{*}{18,885} & 5 & 407 & 17,111 & 2,156 & 2,163 & 21,837 \\
 & \textbf{PTB-XL Sub} & \textbf{Europe} &  & 23 & 407 & 17,111 & 2,156 & 2,163 & 21,837 \\
 & \textbf{PTB-XL Form} & \textbf{Europe} &  & 19 & 12,849 & 7,202 & 904 & 882 & 21,837 \\
 & \textbf{PTB-XL Rhythm} & \textbf{Europe} &  & 12 & 771 & 16,853 & 2,109 & 2,103 & 21,837 \\
 & \textbf{CPSC2018} & \textbf{Asia} & Not opened & 9 & 420 & 4,520 & 646 & 1,291 & 6,877 \\ \bottomrule
\end{tabular}
\end{table*}

CREMA does not achieve the lowest error in any single category, but consistently performs well across all components. As a result, it exhibits the lowest average total error (0.2289), reflecting its balanced learning between representation expressiveness and generalization capacity. This highlights CREMA’s \textbf{stability}, \textbf{versatility}, and suitability for clinical deployment where robustness across data conditions is essential. These findings align with the trends observed in Section~\ref{sec:clinical_eval} and further reinforce CREMA’s reliability as a general-purpose ECG foundation model.

\subsection{Architectural Flexibility}
\begin{figure}[!h]
    \centering
    \begin{subfigure}[]{.8\linewidth}
        \includegraphics[width=1\linewidth]{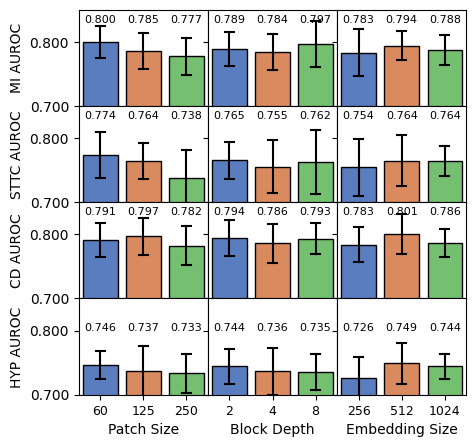}
        \caption{Impact on linear probing performance}
        \label{fig:impact_arc_lp}
    \end{subfigure}
    \begin{subfigure}[]{.8\linewidth}
        \includegraphics[width=1\linewidth]{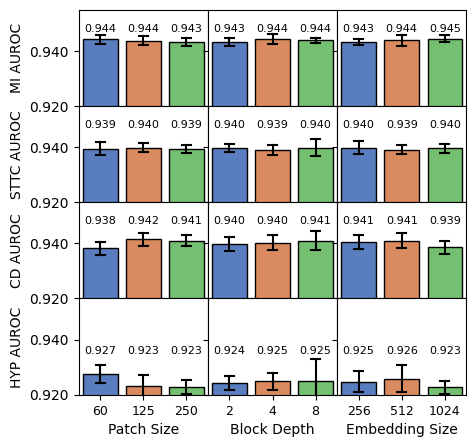}
        \caption{Impact on fine-tuning performance.}
        \label{fig:fine_ar_design}
    \end{subfigure}
    \caption{Impact of the architectural designs.}
    \label{fig:fine_ar_design}
\end{figure}

\begin{table}[!h]
\footnotesize
\centering
\caption{Difference in linear probing performance according to architecture}
\label{tab:differ_lp_perform}
\begin{tabular}{@{}ccrrrr@{}}
\toprule
\textbf{} &
  \textbf{\begin{tabular}[c]{@{}c@{}}Validation\\ Method\end{tabular}} &
  \multicolumn{1}{c}{\textbf{MI}} &
  \multicolumn{1}{c}{\textbf{STTC}} &
  \multicolumn{1}{c}{\textbf{CD}} &
  \multicolumn{1}{c}{\textbf{HYP}} \\ \midrule
\multirow{2}{*}{\textbf{\begin{tabular}[c]{@{}c@{}}Patch\\ Size\\ (60, 125, 250)\end{tabular}}} &
  \textbf{\begin{tabular}[c]{@{}c@{}}ANOVA\\ (F-statistics)\end{tabular}} &
  \begin{tabular}[c]{@{}r@{}}1.216\\ (p \textgreater 0.05)\end{tabular} &
  \begin{tabular}[c]{@{}r@{}}1.774\\ (p \textgreater 0.05)\end{tabular} &
  \begin{tabular}[c]{@{}r@{}}0.468\\ (p \textgreater 0.05)\end{tabular} &
  \begin{tabular}[c]{@{}r@{}}0.323\\ (p \textgreater 0.05)\end{tabular} \\
 &
  \textbf{\begin{tabular}[c]{@{}c@{}}Kruskal-Wallis\\ (H-statistic)\end{tabular}} &
  \begin{tabular}[c]{@{}r@{}}1.918\\ (p \textgreater 0.05)\end{tabular} &
  \begin{tabular}[c]{@{}r@{}}2.725\\ (p \textgreater 0.05)\end{tabular} &
  \begin{tabular}[c]{@{}r@{}}0.730\\ (p \textgreater 0.05)\end{tabular} &
  \begin{tabular}[c]{@{}r@{}}0.940\\ (p \textgreater 0.05)\end{tabular} \\ \midrule
\multirow{2}{*}{\textbf{\begin{tabular}[c]{@{}c@{}}Depth\\ (2, 4, 8)\end{tabular}}} &
  \textbf{\begin{tabular}[c]{@{}c@{}}ANOVA\\ (F-statistics)\end{tabular}} &
  \begin{tabular}[c]{@{}r@{}}0.281\\ (p \textgreater 0.05)\end{tabular} &
  \begin{tabular}[c]{@{}r@{}}0.154\\ (p \textgreater 0.05)\end{tabular} &
  \begin{tabular}[c]{@{}r@{}}0.198\\ (p \textgreater 0.05)\end{tabular} &
  \begin{tabular}[c]{@{}r@{}}0.176\\ (p \textgreater 0.05)\end{tabular} \\ 
 &
  \textbf{\begin{tabular}[c]{@{}c@{}}Kruskal-Wallis\\ (H-statistic)\end{tabular}} &
  \begin{tabular}[c]{@{}r@{}}0.844\\ (p \textgreater 0.05)\end{tabular} &
  \begin{tabular}[c]{@{}r@{}}0.409\\ (p \textgreater 0.05)\end{tabular} &
  \begin{tabular}[c]{@{}r@{}}1.076\\ (p \textgreater 0.05)\end{tabular} &
  \begin{tabular}[c]{@{}r@{}}0.610\\ (p \textgreater 0.05)\end{tabular} \\ \midrule
\multirow{2}{*}{\textbf{\begin{tabular}[c]{@{}c@{}}Embedding\\ Size\\ (256, 512, 1024)\end{tabular}}} &
  \textbf{\begin{tabular}[c]{@{}c@{}}ANOVA\\ (F-statistics)\end{tabular}} &
  \begin{tabular}[c]{@{}r@{}}0.293\\ (p \textgreater 0.05)\end{tabular} &
  \begin{tabular}[c]{@{}r@{}}0.181\\ (p \textgreater 0.05)\end{tabular} &
  \begin{tabular}[c]{@{}r@{}}0.876\\ (p \textgreater 0.05)\end{tabular} &
  \begin{tabular}[c]{@{}r@{}}1.361\\ (p \textgreater 0.05)\end{tabular} \\ 
 &
  \textbf{\begin{tabular}[c]{@{}c@{}}Kruskal-Wallis\\ (H-statistic)\end{tabular}} &
  \begin{tabular}[c]{@{}r@{}}1.001\\ (p \textgreater 0.05)\end{tabular} &
  \begin{tabular}[c]{@{}r@{}}0.312\\ (p \textgreater 0.05)\end{tabular} &
  \begin{tabular}[c]{@{}r@{}}2.061\\ (p \textgreater 0.05)\end{tabular} &
  \begin{tabular}[c]{@{}r@{}}2.256\\ (p \textgreater 0.05)\end{tabular} \\ \bottomrule
\end{tabular}
\end{table}

\begin{table}[!h]
\footnotesize
\centering
\caption{Difference in fine-tuning performance according to architectural design.}
\label{tab:differ_perform}
\begin{tabular}{@{}ccrrrr@{}}
\toprule
\textbf{} &
  \textbf{\begin{tabular}[c]{@{}c@{}}Validation\end{tabular}} &
  \multicolumn{1}{c}{\textbf{MI}} &
  \multicolumn{1}{c}{\textbf{STTC}} &
  \multicolumn{1}{c}{\textbf{CD}} &
  \multicolumn{1}{c}{\textbf{HYP}} \\ \midrule
\multirow{2}{*}{\textbf{\begin{tabular}[c]{@{}c@{}}Patch\\ Size\\ (60, 125, 250)\end{tabular}}} &
  \textbf{\begin{tabular}[c]{@{}c@{}}ANOVA\\ (F-statistics)\end{tabular}} &
  \begin{tabular}[c]{@{}r@{}}0.590\\ (p \textgreater 0.05)\end{tabular} &
  \begin{tabular}[c]{@{}r@{}}0.188\\ (p \textgreater 0.05)\end{tabular} &
  \begin{tabular}[c]{@{}r@{}}4.702\\ (p \textless 0.05)\end{tabular} &
  \begin{tabular}[c]{@{}r@{}}4.562\\ (p \textless 0.05)\end{tabular} \\
 &
  \textbf{\begin{tabular}[c]{@{}c@{}}Kruskal-Wallis\\ (H-statistic)\end{tabular}} &
  \begin{tabular}[c]{@{}r@{}}0.823\\ (p \textgreater 0.05)\end{tabular} &
  \begin{tabular}[c]{@{}r@{}}0.975\\ (p \textgreater 0.05)\end{tabular} &
  \begin{tabular}[c]{@{}r@{}}6.061\\ (p \textless 0.05)\end{tabular} &
  \begin{tabular}[c]{@{}r@{}}8.640\\ (p \textless 0.05)\end{tabular} \\ \midrule
\multirow{2}{*}{\textbf{\begin{tabular}[c]{@{}c@{}}Depth\\ (2, 4, 8)\end{tabular}}} &
  \textbf{\begin{tabular}[c]{@{}c@{}}ANOVA\\ (F-statistics)\end{tabular}} &
  \begin{tabular}[c]{@{}r@{}}0.917\\ (p \textgreater 0.05)\end{tabular} &
  \begin{tabular}[c]{@{}r@{}}0.423\\ (p \textgreater 0.05)\end{tabular} &
  \begin{tabular}[c]{@{}r@{}}0.315\\ (p \textgreater 0.05)\end{tabular} &
  \begin{tabular}[c]{@{}r@{}}0.093\\ (p \textgreater 0.05)\end{tabular} \\ 
 &
  \textbf{\begin{tabular}[c]{@{}c@{}}Kruskal-Wallis\\ (H-statistic)\end{tabular}} &
  \begin{tabular}[c]{@{}r@{}}1.044\\ (p \textgreater 0.05)\end{tabular} &
  \begin{tabular}[c]{@{}r@{}}1.007\\ (p \textgreater 0.05)\end{tabular} &
  \begin{tabular}[c]{@{}r@{}}0.624\\ (p \textgreater 0.05)\end{tabular} &
  \begin{tabular}[c]{@{}r@{}}0.453\\ (p \textgreater 0.05)\end{tabular} \\ \midrule
\multirow{2}{*}{\textbf{\begin{tabular}[c]{@{}c@{}}Embedding\\ Size\\ (256, 512, 1024)\end{tabular}}} &
  \textbf{\begin{tabular}[c]{@{}c@{}}ANOVA\\ (F-statistics)\end{tabular}} &
  \begin{tabular}[c]{@{}r@{}}1.390\\ (p \textgreater 0.05)\end{tabular} &
  \begin{tabular}[c]{@{}r@{}}0.337\\ (p \textgreater 0.05)\end{tabular} &
  \begin{tabular}[c]{@{}r@{}}1.694\\ (p \textgreater 0.05)\end{tabular} &
  \begin{tabular}[c]{@{}r@{}}1.075\\ (p \textgreater 0.05)\end{tabular} \\ 
 &
  \textbf{\begin{tabular}[c]{@{}c@{}}Kruskal-Wallis\\ (H-statistic)\end{tabular}} &
  \begin{tabular}[c]{@{}r@{}}3.647\\ (p \textgreater 0.05)\end{tabular} &
  \begin{tabular}[c]{@{}r@{}}0.600\\ (p \textgreater 0.05)\end{tabular} &
  \begin{tabular}[c]{@{}r@{}}3.704\\ (p \textgreater 0.05)\end{tabular} &
  \begin{tabular}[c]{@{}r@{}}2.290\\ (p \textgreater 0.05)\end{tabular} \\ \bottomrule
\end{tabular}
\end{table}

This section investigates how architectural parameters—patch size, block depth, and embedding size—affect CREMA’s performance across classification tasks. We trained 27 CREMA models, each representing a unique combination of patch sizes (60, 125, 250), block depths (2, 4, 8), and embedding sizes (256, 512, 1024). Performance was evaluated using both linear probing and fine-tuning across four tasks: MI, STTC, CD, and HYP.

Figure~\ref{fig:fine_ar_design} presents the fine-tuning results, where the x-axis denotes architectural configurations and the y-axis shows the mean AUROC across the four tasks. Error bars indicate standard deviation, illustrating performance consistency across settings.

To assess statistical significance, we applied ANOVA and Kruskal-Wallis tests, with all p-values exceeding 0.5 (Table~\ref{tab:differ_perform}), indicating no meaningful performance differences attributable to architecture. Linear probing results mirrored this trend.

These findings demonstrate CREMA’s robustness to architectural variations, reducing sensitivity to hyperparameter tuning and supporting its reliability across deployment scenarios.

\section{Data sets} \label{apdx:dataset}
This section describes the various public datasets used in our study, including MIMIC-IV, CODE15, UK Biobank, SaMi-Trop, IKEM, PTB-XL, and CPSC2018. 
From these datasets, we separate the two datasets: the pre-trained dataset and the downstream dataset.
The summaries of each dataset are presented in Table~\ref{tab:summary_dataset}, including demographic information, the number of samples, data split, and specific details about the data collection and characteristics.

\section{Evaluation Metrics} \label{apdx:metrics}


\noindent\textbf{\textit{Area Under the Receiver Operating Characteristic (AUROC)}} The Area Under the Receiver Operating Characteristic (AUROC) curve is a statistical measure that evaluates the performance of binary classification models. AUROC plots the True Positive Rate (TPR) versus the False Positive Rate (FPR) at different threshold settings. It represents the probability of a classifier ranking a randomly chosen positive instance higher than a randomly chosen negative one. An AUC of 1 indicates perfect classification, while an AUC of 0.5 suggests performance equivalent to random guessing. AUROC is useful for evaluating models on imbalanced datasets as it is not influenced by class label distribution.

\noindent\textbf{\textit{Area Under the Precision-Recall Curve (AUPRC)}} The Area Under the Precision-Recall Curve (AUPRC) provides a measure to evaluate binary classification model performance, especially under class imbalance. Unlike AUROC, which plots TPR against FPR, PRC plots Precision (true positives to all predicted positives) against Recall (equivalent to TPR). A higher AUPRC value represents better performance in distinguishing between classes under imbalanced class distributions.

\end{document}